\documentclass{article}

\usepackage[final,nonatbib]{tackling_climate_workshop_style}

\usepackage[utf8]{inputenc} 
\usepackage[T1]{fontenc}    
\usepackage{hyperref}       
\usepackage{url}            
\usepackage{booktabs}       
\usepackage{amsfonts}       
\usepackage{nicefrac}       
\usepackage{microtype}      
\usepackage{multirow}
\usepackage{subfigure}
\usepackage[flushleft]{threeparttable}
\usepackage{xcolor}
\usepackage{graphicx}
\usepackage{tabularx}
\usepackage{amsmath,amssymb,amsfonts}
\usepackage{algorithmic}
\usepackage{textcomp}
\usepackage[notextcomp]{stix}
\usepackage{amsmath}
\usepackage{indentfirst}
\usepackage{siunitx}

\definecolor{s5}{RGB}{255, 234, 221}
\definecolor{5t10}{RGB}{252, 174, 174}
\definecolor{10t20}{RGB}{255, 137, 137}
\definecolor{l20}{RGB}{255, 102, 102}
\definecolor{best1}{RGB}{122,189,129}
\definecolor{best2}{RGB}{149,198,132}
\definecolor{best3}{RGB}{208,235,140}
\definecolor{best4}{RGB}{239,230,144}
\definecolor{worst1}{RGB}{250,225,143}
\definecolor{worst2}{RGB}{241,179,130}
\definecolor{worst3}{RGB}{237,159,124}
\definecolor{worst4}{RGB}{231,114,111}

\title{DeepEn2023: Energy Datasets for Edge Artificial Intelligence}
\author{%
  Xiaolong Tu \\
  Department of Computer Science\\
  Georgia State University \\
  Atlanta, GA 30302 \\ 
  \texttt{xtu1@student.gsu.edu} \\
  \And
  Anik Mallik \\
  Department of Electrical and Computer Engineering\\
  The University of North Carolina at Charlotte \\
  Charlotte, NC 28223\\
  \texttt{amallik@uncc.edu} \\
  \AND
  Haoxin Wang\\
  Department of Computer Science\\
  Georgia State University \\
  Atlanta, GA 30302 \\ 
  \texttt{haoxinwang@gsu.edu} \\
  \And
  Jiang Xie \\
  Department of Electrical and Computer Engineering\\
  The University of North Carolina at Charlotte \\
  Charlotte, NC 28223\\
  \texttt{linda.xie@uncc.edu} \\
}

\begin{document}

\maketitle
\begin{abstract}
Climate change poses one of the most significant challenges to humanity. 
As a result of these climatic changes, the frequency of weather, climate, and water-related disasters has multiplied fivefold over the past 50 years, resulting in over $2$ million deaths and losses exceeding \$3.64 trillion USD.
Leveraging AI-powered technologies for sustainable development and combating climate change is a promising avenue. Numerous significant publications are dedicated to using AI to improve renewable energy forecasting, enhance waste management, and monitor environmental changes in real time. However, very few research studies focus on making AI itself environmentally sustainable. 
This oversight regarding the sustainability of AI within the field might be attributed to a mindset gap and the absence of comprehensive energy datasets. In addition, with the ubiquity of edge AI systems and applications, especially on-device learning, there is a pressing need to measure, analyze, and optimize their environmental sustainability, such as energy efficiency.
To this end, in this paper, we propose large-scale energy datasets for edge AI, named DeepEn2023, covering a wide range of kernels, state-of-the-art deep neural network models, and popular edge AI applications.
We anticipate that DeepEn2023 will improve transparency in sustainability in on-device deep learning across a range of edge AI systems and applications. For more information, including access to the dataset and code, please visit https://amai-gsu.github.io/DeepEn2023.

\end{abstract}

\section{Introduction}
Environmentally-sustainable AI refers to the design and use of artificial intelligence (AI) and machine learning (ML) technologies to tackle environmental issues and advance sustainability \cite{wu2022sustainable}, which is a two-sided research area: AI for sustainability and sustainability of AI \cite{van2021sustainable,vinuesa2020role}. While there is growing interest in using AI to achieve the Sustainable Development Goals (SDGs) \cite{SDG} related to climate change, research addressing the environmental impact of AI itself remains limited \cite{you2023zeus} \cite{wang2021asymo} \cite{chen2022deepperform} \cite{cai2021towards}. 
For instance, a sophisticated AI-empowered Internet of Things (IoT) system can be deployed to monitor and predict the total carbon emissions of a building or factory, aligning with the objective of AI for sustainability.
However, this raises new questions: \textit{How much carbon does this AI system emit? How sustainable is the AI system itself?}

On-device learning on edge devices, such as smartphones, IoT devices, and connected vehicles, is increasingly prevalent for model personalization and enhanced data privacy, yet its impact in terms of carbon emission is often overlooked \cite{wu2022sustainable, wu2019machine,savazzi2021framework}. This oversight might be attributed to the typically modest power consumption and carbon footprint of individual edge devices. However, when considering the immense proliferation of these AI-empowered devices worldwide, their cumulative carbon footprint would be substantial and cannot be overlooked.
For instance, consider a scenario where an individual uses AI-powered applications on their smartphone for one hour every day. The average power consumption of a smartphone is $3$W \cite{wang2020energy}. 
With $6.4$ billion smartphone connections reported in $2022$ \cite{GSMA2023}, the cumulative energy consumption of these smartphones amounts to $19,200$ MWh per day. Based on the U.S. electricity generation carbon intensity of $371.2$kg of carbon per MWh \cite{USelectricitygeneration}, the estimated daily carbon emissions from these smartphones would be $7127.04$ metric tons. For comparison, this is equivalent to the annual carbon footprint of $1,848$ gasoline-powered passenger vehicles. \cite{EPA}.
Therefore, to understand and evaluate the sustainability of AI systems, especially edge AI systems, we have developed three large-scale energy consumption datasets: \textit{kernel-level, model-level, and application-level}.
We hope our energy datasets, named \textit{DeepEn2023}, will encourage both the research community and end-users to prioritize sustainability in on-device learning and edge AI, a principle that drives our research.

 
\section{Energy Measurement Platform}
\begin{figure*}[t]
\centering
{\includegraphics[width=1.0\textwidth]{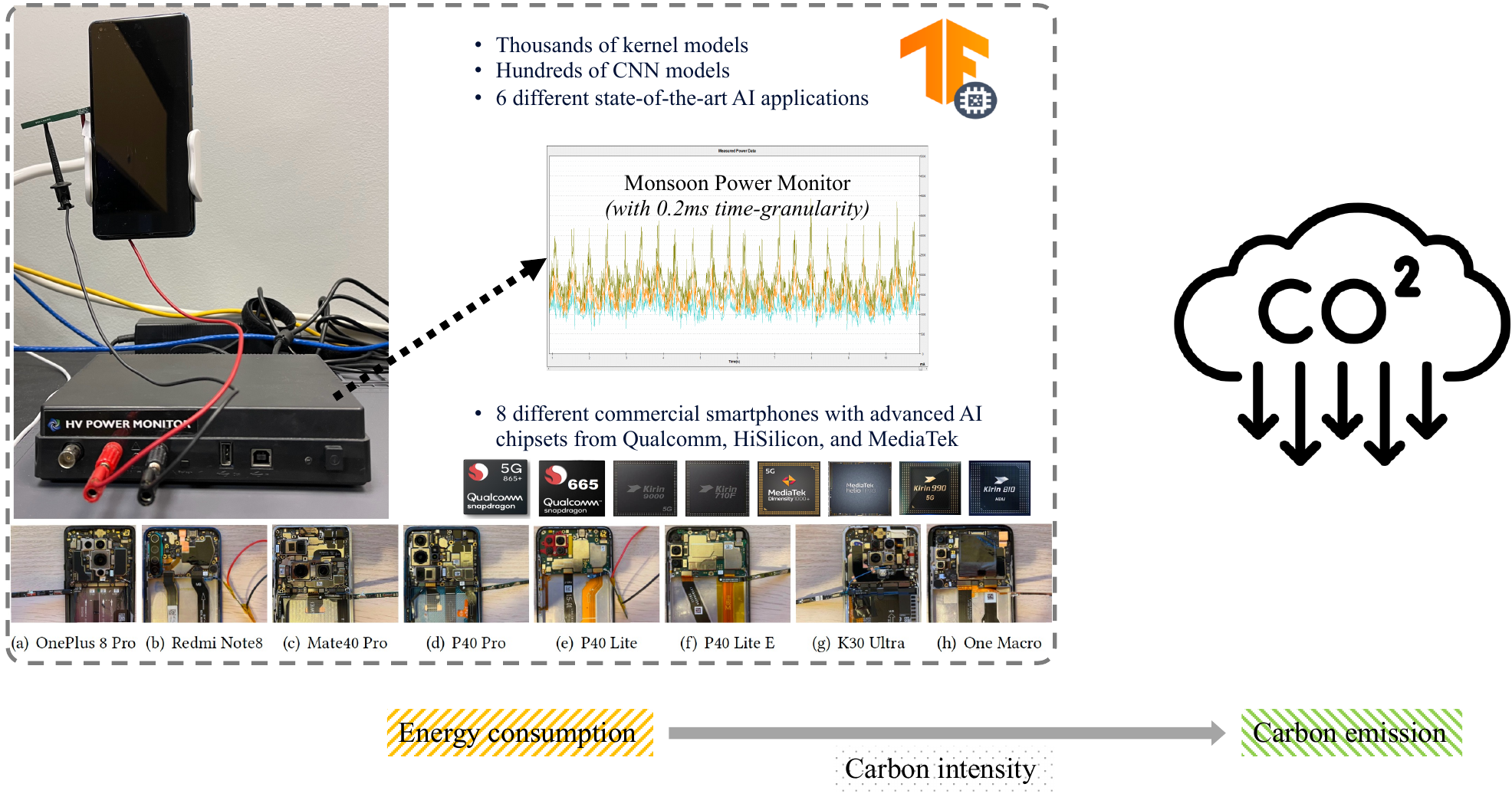}}
\caption{Power Measurement Platform utilizing the Monsoon Power Monitor to capture energy consumption data. Then, carbon intensity are used to convert this energy data into carbon emission estimates.}
\label{fig:platform}
\end{figure*}

We developed an energy measurement platform employing the Monsoon Power Monitor to capture power consumption data during model execution. This power data, combined with inference latency, is used to generate energy datasets. The Monsoon Power Monitor is selected for its millisecond-level data granularity. Since most DNN model latencies, typically between 10 to 200 ms on mobile CPUs, can be significantly decreased to 1 to 50 ms on mobile GPUs. Compared to built-in smartphone sensors, the Monsoon provides more accurate and detailed power consumption data, especially for models running on edge devices. Fig.\ref{fig:platform} illustrates the power measurement platform we have implemented. We connected battery-removed smartphones to the power monitor using power cables. Then use Monsoon power monitor to power on the devices and measure power consumption during model execution with a granularity of up to 0.2 ms. We generated thousands of TensorFlow Lite models across various levels and executed them on different hardware platforms to create a comprehensive dataset.

For our study, we selected eight modern edge devices featuring eight different mobile SoCs, including at least one high-end and one mid-range SoC from leading chipset vendors such as Qualcomm, HiSilicon, and MediaTek. These SoCs have been chosen for their status as representative and advanced mobile AI silicon widely used in the last two years.

\section{DeepEn2023: Energy Consumption Dataset}
In this section, we provide details of our datasets and how it contributes to understanding the energy consumption and carbon emissions of edge AI systems. We have generated comprehensive datasets for typical kernels, models and applications across various configurations. We also discuss how each dataset can facilitate research efforts aimed at accessing the adverse impact of AI carbon emissions on global climate change.

\subsection{Kernel-level Energy Consumption Dataset}
\label{sc:kernelDataset}
Kernels constitute the fundamental units of execution in deep learning frameworks, with their types and configuration parameters significantly influencing the energy consumption during DNN model executions.
In Table \ref{tb:ker_data} we list nine typical kernels that are present in almost all CNN models, with the energy consumption and the carbon emission range for different configurations. The primary configurations include input height and width ($HW$)\footnote{In CNN models, input height usually is equal to input width.}, input channel number ($C_{in}$), output channel number ($C_{out}$), kernel size ($KS$), and stride ($S$). Here are the key observations : 
1) Energy consumption varies significantly for same kernel with different configurations on CPU and GPU. 
2) Different configuration parameters have varying impacts for the kernels energy consumption 
3) \texttt{conv$\doubleplus$bn$\doubleplus$relu} kernels typically consume more energy than other kernel types. 
4) Across almost all the kernels, GPU exhibit better energy efficiency under same configurations. 
Studying the impact of kernel configurations on energy consumption lays the foundation for a comprehensive understanding of energy usage during DNN model executions on edge devices. This emphasizes the importance of adaptive configuration selecting, in enhancing the energy efficiency of DNN models and how it can benefit researchers working toward carbon netural goal.

\begin{table*}[t]
\centering
\caption{Measured kernels per device in our kernel-level dataset.}
\resizebox{\textwidth}{!}{
\begin{tabular}{llllllll}
\toprule
\multicolumn{1}{c}{\multirow{3}{*}{Kernels}} & \multicolumn{2}{l}{Energy Consumption (mJ)}                                                                                                                                                  & \multicolumn{2}{l}{Carbon Emission (gCO2eq/kWh)\footnotemark[2]}                                                                                                                                                 & \multicolumn{2}{l}{\# Measured kernels}                                              & \multicolumn{1}{c}{\multirow{3}{*}{Configurations}} \\ 
\cline{2-7}
    & CPU & GPU &  CPU & GPU & \multirow{2}{*}{CPU} &  \multirow{2}{*}{GPU}  &   \\
    & min - max & min - max
    & min - max & min - max &          &     &      \\                              
\hline
\texttt{conv$\doubleplus$bn$\doubleplus$relu}    & 0.002 - 1200.083  & 0.002 - 120.152  &  \num{1.762e-10} - \num{1.057e-4} & \num{1.762e-10} - \num{1.058e-5} &1032&1032&  ($HW, C_{in}, C_{out}, KS, S$)    \\
\texttt{dwconv$\doubleplus$bn$\doubleplus$relu}  & 0.022 - 222.609   & 0.016 - 0.658    &  \num{1.938e-9} - \num{1.961e-5}  & \num{1.409e-9} - \num{5.797e-8} &349&349&    ($HW, C_{in}, KS, S$)    \\ 
\texttt{bn$\doubleplus$relu}                     & 0.002 - 161.334   & 0.001 - 14.594   &  \num{1.762e-10} - \num{1.421e-5} & \num{8.811e-11} - \num{1.285e-6} &100&100&  ($HW, C_{in}$)    \\ 
\texttt{relu}                                    & 0.001 - 141.029   & 0.003 - 6.86     &  \num{8.811e-11} - \num{1.242e-5}  & \num{2.643e-10} - \num{6.044e-7} &46&46&  ($HW, C_{in}$)    \\ 
\texttt{avgpool}                                 & 0.066 - 7.711     & 0.034 - 1.142    &   \num{5.815e-9} - \num{6.794e-7}  & \num{2.995e-9} - \num{1.006e-7} &28&28&  ($HW, C_{in}, KS, S$)    \\ 
\texttt{maxpool}                                 & 0.054 - 7.779     & 0.032 - 1.214    &   \num{4.758e-9} - \num{6.854e-7}   &  \num{2.819e-9} - \num{1.069e-7}&28&28&  ($HW, C_{in}, KS, S$)    \\ 
\texttt{fc}                                      & 0.038 - 94.639    & -                &   \num{3.348e-9} - \num{8.338e-7}  & -             &24&-&  ($C_{in}, C_{out}$)    \\
\texttt{concat}                                  & 0.001 - 42.826    & 0.066 - 3.428    &    \num{8.811e-11} - \num{3.773e-6} &  \num{5.815e-9} - \num{3.020e-7} &142&142&  ($HW, C_{in1}, C_{in2}, C_{in3}, C_{in4}$)    \\ 
\texttt{others}                                  & 0.001 - 132.861   & 0.003 - 10.163   &  \num{8.811e-11} - \num{1.170e-5} &  \num{2.643e-10} - \num{8.954e-7} &98&72&  ($HW, C_{in}$)    \\ 
\bottomrule

\end{tabular}}
\label{tb:ker_data}
\end{table*}
\footnotetext[2]{The unit of measurement typically used for quantifying and comparing carbon emissions is CO2 equivalents.}

To build the dataset, we initially generate a large number of kernels with a variety of types (16 types for CPU and 10 types for GPU) featuring a range of configurations in the \texttt{tflite} format (e.g., $1032$ \texttt{conv$\doubleplus$bn$\doubleplus$relu} and $349$ \texttt{dwconv$\doubleplus$bn$\doubleplus$relu} kernels). These kernel configurations are randomly sampled. The number of sampled configurations for each kernel type hinges on two main factors: its configuration dimension and its impact on the overall energy consumption during DNN executions.
This dataset provides researchers with detailed insights into how energy is consumed within models and which configurations or parameters affect kernel energy efficiency. Researchers can use this dataset to adapt configurations with best erergy efficiency on edge devices, consequently reducing carbon emissions.

\subsection{Model-level Energy Consumption Dataset}
\begin{figure*}[t]
\centering
\subfigure[Mobile CPU]
{\includegraphics[width=0.495\textwidth]{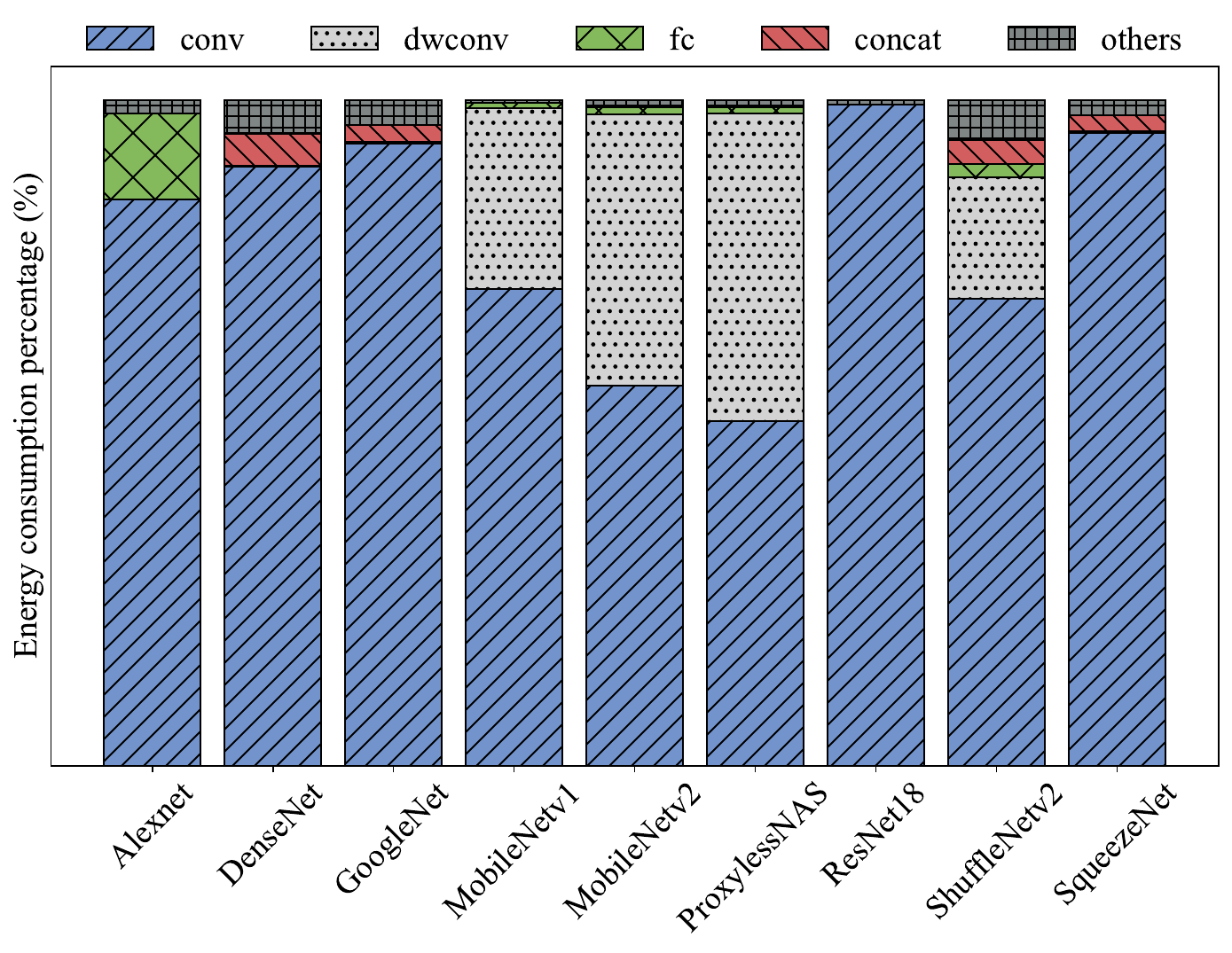}}\label{fig:rescpu}
\subfigure[Mobile GPU]
{\includegraphics[width=0.495\textwidth]{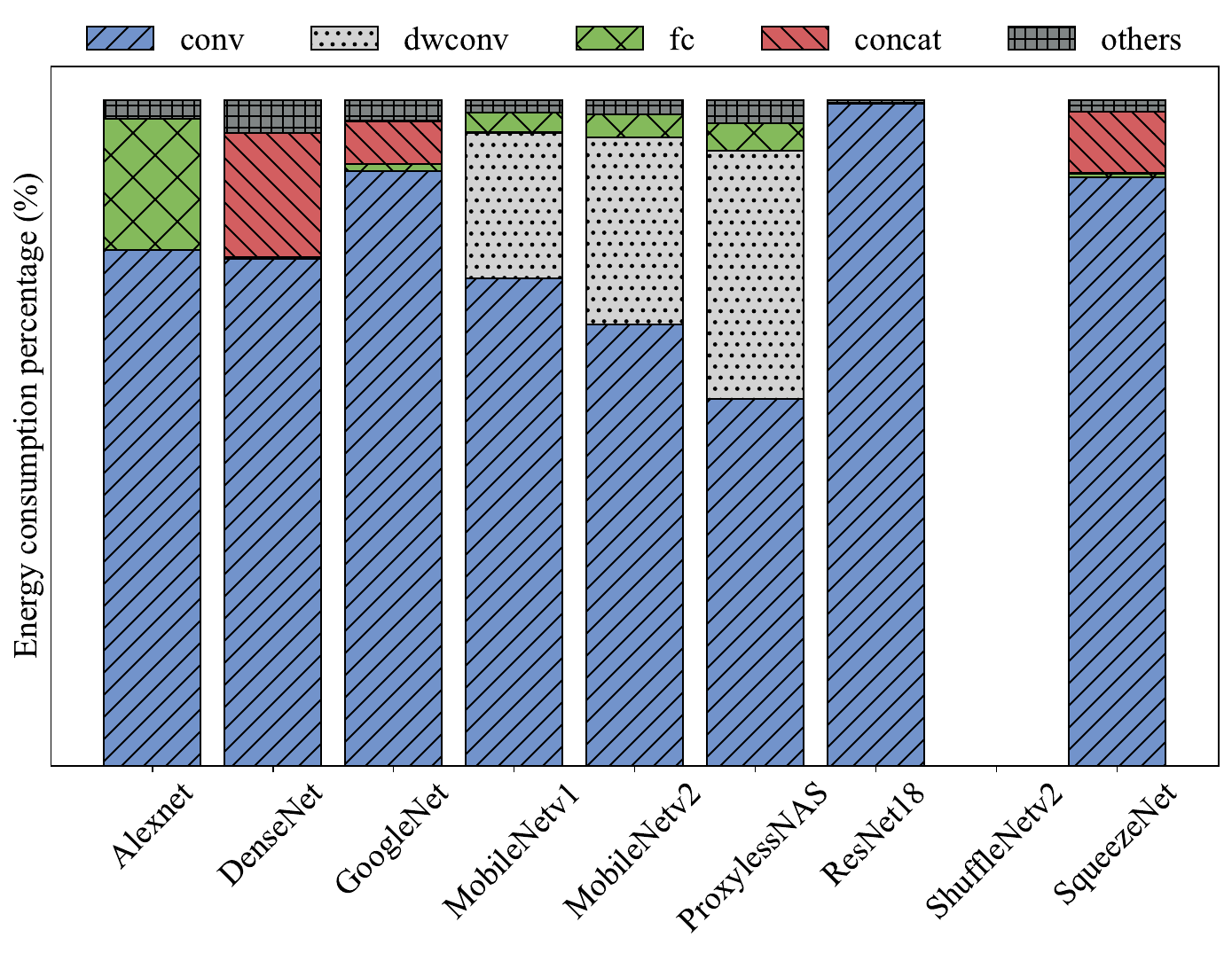}}\label{fig:resgpu}
\caption{DNN model energy consumption percentage breakdown. The top four most energy-consuming kernel types are \texttt{conv$\doubleplus$bn$\doubleplus$relu} (\texttt{conv}), \texttt{dwconv$\doubleplus$bn$\doubleplus$relu} (\texttt{dwconv}), \texttt{fc}, and \texttt{concat}.}
\label{fig:preres}   
\end{figure*}

We also introduce our model-level energy dataset, which collects nine state-of-the-art DNN models. These models represent a mix of both manually-designed and NAS-derived models, each with distinct kernel types and configurations. For each model, we generate $50$ variants for conducting power and energy measurements by re-sampling the $C_{out}$ and $KS$ for each layer. Specifically, we randomly sample the new output channel number from a range of $20\%$ to $180\%$ of the original $C_{out}$, while the $KS$ is sampled from the set of values: $\{1, 3, 5, 7, 9\}$. Generally, running these models on mobile GPUs results in an energy consumption reduction of approximately $49\%$ to $79\%$, compared to the execution on mobile CPUs. Fig. \ref{fig:preres} presents the energy consumption breakdown of individual models by kernel types.  
The four kernel types that consume the most energy are \texttt{conv$\doubleplus$bn$\doubleplus$relu}, \texttt{dwconv$\doubleplus$bn$\doubleplus$relu}, \texttt{fc}, and \texttt{concat}. They account for $79.27\%$, $14.79\%$, $2.03\%$, and $1.5\%$ of the total model energy consumption on the mobile CPU, respectively. On the mobile GPU, these kernels represent $78.17\%$, $10.91\%$, $4.01\%$, and $4.28\%$ of the total model energy consumption.
Furthermore, in most models, \texttt{conv$\doubleplus$bn$\doubleplus$relu} and \texttt{dwconv$\doubleplus$bn$\doubleplus$relu} account for the main energy percentages. On average, \texttt{conv$\doubleplus$bn$\doubleplus$relu} and \texttt{dwconv$\doubleplus$bn$\doubleplus$relu} take $93.97\%$ and $87.74\%$ of the total model energy consumption on the mobile CPU and GPU, respectively. 
With this model-level energy consumption dataset, researchers can visually see the energy consumption of different models on various platforms, helping them choose he most energy-efficient models according to their needs.

\subsection{Application-level Energy Consumption Dataset}
\begin{figure*}[t]
\centering
\subfigure[End-to-end processing pipeline for object detection and image classification]
{\includegraphics[width=0.495\textwidth]{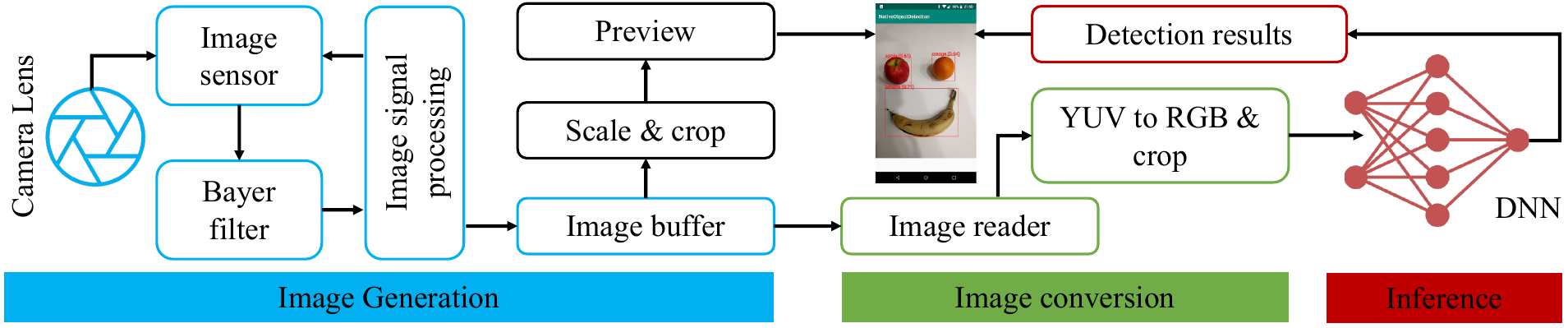}}\label{fig:pipline}
\subfigure[Energy consumption percentage breakdown]
{\includegraphics[width=0.48\textwidth]{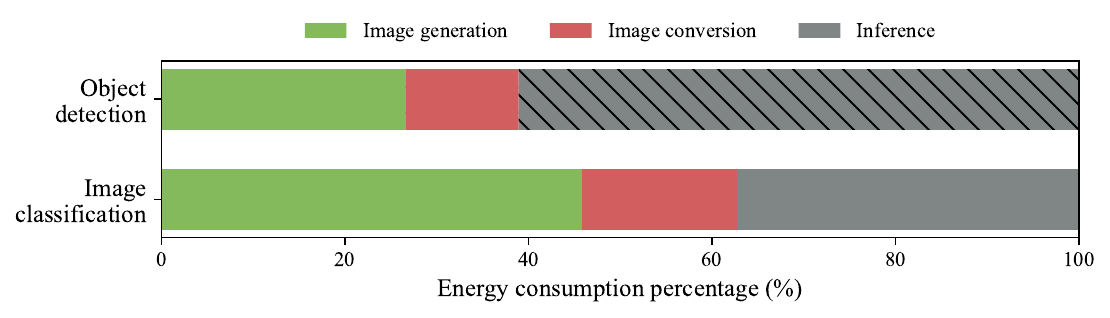}}\label{fig:breakdown}
\caption{End-to-end energy consumption breakdown for object detection and image classification based on our application-level dataset.}
\label{fig:e2e}   
\end{figure*}

The kernel- and model-level datasets can be beneficial for researchers and developers in understanding, modelling, and optimizing power and energy efficiency of DNN executions. However, the energy efficiency of applications on edge devices has a more direct impact on carbon emissions. 
To adress this, we create an application-level dataset, which uncovers the end-to-end energy consumption of six popular edge AI applications, covering three main categories: vision-based (object detection, image classification, super resolution, and image segmentation), NLP-based (natural language question answering), and voice-based applications (speech recognition).
As shown in Table \ref{tab:appdata}, we measure the power and energy consumption of each application with multiple reference DNN models that operate under four distinct computational settings, including CPU with a single thread, CPU with four threads, GPU delegate, and the NNAPI delegate.
The dataset can serve as a resource for exploring the energy consumption distribution throughout the end-to-end processing pipeline of an edge AI application.
For example, we can use the dataset to examine the energy consumed in generating image frames, converting these frames from YUV to RGB, and conducting DNN inference within an object detection application. 
It demonstrates that our application-level dataset can provide interpretable observations for comprehending who is the primary energy consumer in the end-to-end edge AI application.
Fig. \ref{fig:e2e} depicts the energy consumption breakdown based on the processing phases in the object detection. It demonstrates that our application-level dataset can provide interpretable observations for comprehending who is the primary energy consumer in the end-to-end edge AI application.

\begin{table*}
\caption{Measured edge AI applications per device in our application-level dataset.}
  \label{tab:appdata}
\resizebox{\textwidth}{!}
{
\begin{tabular}{llllllll}

\toprule
\multicolumn{1}{c}{}                                    & \multicolumn{1}{c}{}                                       & \multicolumn{1}{c}{}                                                            & \multicolumn{4}{c}{Delegate}                                                                                            & \multicolumn{1}{c}{}                                                                                     \\ \cline{4-7}
\multicolumn{1}{c}{\multirow{-2}{*}{Category}} & \multicolumn{1}{c}{\multirow{-2}{*}{Application}} & \multicolumn{1}{c}{\multirow{-2}{*}{Reference DNN models}} & \multicolumn{1}{l}{CPU1} & \multicolumn{1}{l}{CPU4} & \multicolumn{1}{l}{GPU}                     & NNAPI                      & \multicolumn{1}{c}{\multirow{-2}{*}{\begin{tabular}[c]{@{}c@{}}Model size\\ (MB)\end{tabular}}} \\ \hline
                                                          &                                                                                                              & MobileNetv2, FP32, 300 × 300 pixels            & \multicolumn{1}{l} {\color{green} \checkmark}  & \multicolumn{1}{l} {\color{green} \checkmark}  & \multicolumn{1}{l}{}                        &  {\color{green} \checkmark}                         & 24.2                                                                                                      \\ \cline{3-8} 
                                                          &                                                                                                            & MobileNetv2, INT8, 300 × 300 pixels            & \multicolumn{1}{l} {\color{green} \checkmark}  & \multicolumn{1}{l} {\color{green} \checkmark}  & \multicolumn{1}{l}{}                        &  {\color{green} \checkmark}                         & 6.9                                                                                                       \\ \cline{3-8} 
                                                          &                                                                                                        & MobileNetv2, FP32, 640 × 640 pixels   & \multicolumn{1}{l} {\color{green} \checkmark}  & \multicolumn{1}{l} {\color{green} \checkmark}  & \multicolumn{1}{l}{}                        &  {\color{green} \checkmark}                         & 12.3                                                                                                      \\ \cline{3-8} 
                                                          & \multirow{-4}{*}{Image detection}                                                                             & MobileNetv2, INT8, 640 × 640 pixels   & \multicolumn{1}{l} {\color{green} \checkmark}  & \multicolumn{1}{l} {\color{green} \checkmark}  & \multicolumn{1}{l}{}                        &  {\color{green} \checkmark}                         & 4.5                                                                                                       \\ \cline{2-8} 
                                                          &                                                                                                              & EfficientNet, FP32, 224 × 224 pixels                & \multicolumn{1}{l} {\color{green} \checkmark}  & \multicolumn{1}{l} {\color{green} \checkmark}  & \multicolumn{1}{l} {\color{green} \checkmark}                     &  {\color{green} \checkmark}                         & 18.6                                                                                                      \\ \cline{3-8} 
                                                          &                                                                                                               & EfficientNet, INT8, 224 × 224 pixels                & \multicolumn{1}{l} {\color{green} \checkmark}  & \multicolumn{1}{l} {\color{green} \checkmark}  & \multicolumn{1}{l}{}                        &  {\color{green} \checkmark}                         & 5.4                                                                                                       \\ \cline{3-8} 
                                                          &                                                                                                             & MobileNetv1, FP32, 224 × 224 pixels                & \multicolumn{1}{l} {\color{green} \checkmark}  & \multicolumn{1}{l} {\color{green} \checkmark}  & \multicolumn{1}{l} {\color{green} \checkmark}                     &  {\color{green} \checkmark}                         & 4.3                                                                                                       \\ \cline{3-8} 
                                                          & \multirow{-4}{*}{Image classification}                                                                       & MobileNetv1, INT8, 224 × 224 pixels                & \multicolumn{1}{l} {\color{green} \checkmark}  & \multicolumn{1}{l} {\color{green} \checkmark}  & \multicolumn{1}{l}{}                        &  {\color{green} \checkmark}                         & 16.9                                                                                                      \\ \cline{2-8} 
                                                          & Super resolution                                                                                             & ESRGAN , FP32, 50 × 50 pixels                         & \multicolumn{1}{l} {\color{green} \checkmark}  & \multicolumn{1}{l}{}     & \multicolumn{1}{l} {\color{green} \checkmark}                     &                            & 5                                                                                                         \\ \cline{2-8} 
\multirow{-10}{*}{Vision-based}            & Image segmentation                                                                                           & DeepLabv3 , FP32, 257 × 257 pixels                  & \multicolumn{1}{l}{}     & \multicolumn{1}{l} {\color{green} \checkmark}  & \multicolumn{1}{l}{}                        &                            & 2.8                                                                                                       \\ \hline
NLP-based                                  & Natural language question answering                                                                                                   & MobileBERT , FP32                                   & \multicolumn{1}{l} {\color{green} \checkmark}  & \multicolumn{1}{l} {\color{green} \checkmark}  & \multicolumn{1}{l}{}                        &  {\color{green} \checkmark}                         & 100.7                                                                                                     \\ \hline
Voice-based                                & Speech recognition                                                                                         & Conv-Actions-Frozen , FP32                           & \multicolumn{1}{l} {\color{green} \checkmark}  & \multicolumn{1}{l} {\color{green} \checkmark}  & \multicolumn{1}{l}{{\color[HTML]{FE0000} }} & {\color{green}  {\color{green} \checkmark} } & 3.8                                                                                                       \\
\bottomrule
\end{tabular}
}    
\end{table*}

\subsection{Beneficial For Global Climate Change}
These three datasets can contribute to addressing global climate change from different perspectives. For example, the kernel-level dataset can assist researchers in identifying the most energy-efficient kernel configurations and parameters, finding the balance between computing performance and carbon emissions. We have used our dataset to train a random forest model to predict the energy consumption and carbon emissions of unseen models, and the accuracy is quite promising \cite{tu2023energy}. The model-level dataset aids researchers in discovering the most energy-efficient models based on various deployment requirements. For instance, Model A deployed on a CPU may exhibit better energy efficiency than Model B with the same accuracy for image classification. The application-level dataset provides researchers with insights into the end-to-end energy consumption of an application on edge devices, enabling them to implement more comprehensive measures to reduce energy consumption.

\section{Conclusion}
In this paper, we present our energy consumption datasets, DeepEn2023, from kernel-level, model-level, and application-level to facilitate research and development aimed at improving the energy efficiency and reducing the carbon emissions of AI applications on diverse edge devices. These datasets are valuable resources and tools for researchers and community to design energy-efficiency AI systems with fewer greenhouse gas emissions, thus contributing to the global climate change mitigation. We hope DeepEn2023 can help shift the mindset of both end-users and the research community towards sustainable edge AI, a principle that drives our research.

\acksection{This work was supported by the US National Science Foundation (NSF) under Grant No. 1910667, 1910891, and 2025284.}

\medskip

\clearpage

\bibliographystyle{unsrt}
\bibliography{reference}


\end{document}